\documentclass[letterpaper, 10 pt, conference]{ieeeconf} 
\IEEEoverridecommandlockouts                             
\usepackage{graphicx} 
\usepackage{amsmath} 
\usepackage{amssymb}  
\usepackage{booktabs}
\usepackage{multirow}
\usepackage{float}
\usepackage{tabularx}
\usepackage{pifont}
\usepackage[most]{tcolorbox}
\usepackage{wrapfig}
\usepackage{bm}
\usepackage{xcolor}
\usepackage{adjustbox}
\usepackage{arydshln}
\usepackage{tikz}
\usepackage{placeins}
\usepackage{caption}
\usepackage{bbold}
\usepackage{mathtools}

\newcommand{\dashrule}[1][black]{%
  \color{#1}\rule[\dimexpr.5ex-.2pt]{4pt}{.4pt}\xleaders\hbox{\rule{4pt}{0pt}\rule[\dimexpr.5ex-.2pt]{4pt}{.4pt}}\hfill\kern0pt%
}

\newcommand{\etal}{\emph{et al.}}

\title{\LARGE \bf Distilling Knowledge for Short-to-Long Term Trajectory Prediction}
\author{Sourav Das$^{1}$, Guglielmo Camporese$^{2}$, Shaokang Cheng$^{3}$, and Lamberto Ballan$^{1}$
\thanks{
    $^{1}$Sourav Das and Lamberto Ballan are with the Department of Mathematics ``Tullio Levi-Civita'', University of Padova, Italy. $^{2}$Guglielmo Camporese recently earned his Ph.D. from the same department and he is currently a Research Engineer at Disney Research, Zurich. $^{3}$S. Cheng is with the School of Automation, Northwestern Polytechnical University, China.
}
}

\begin{document}
\maketitle
\thispagestyle{empty}
\pagestyle{empty}


\begin{abstract}
Long-term trajectory forecasting is an important and challenging problem in the fields of computer vision, machine learning, and robotics. One fundamental difficulty stands in the evolution of the trajectory that becomes more and more uncertain and unpredictable as the time horizon grows, subsequently increasing the complexity of the problem. To overcome this issue, in this paper, we propose \textit{Di-Long}, a new method that employs the distillation of a short-term trajectory model forecaster that guides a student network for long-term trajectory prediction during the training process. 
Given a total sequence length that comprehends the allowed observation for the student network and the complementary target sequence, we let the student and the teacher solve two different related tasks defined over the same full trajectory: the student observes a short sequence and predicts a long trajectory, whereas the teacher observes a longer sequence and predicts the remaining short target trajectory. 
The teacher's task is less uncertain, and we use its accurate predictions to guide the student through our knowledge distillation framework, reducing long-term future uncertainty.
Our experiments show that our proposed Di-Long method is effective for long-term forecasting and achieves state-of-the-art performance on the Intersection Drone Dataset (inD) and the Stanford Drone Dataset (SDD).
\end{abstract}


\vspace{2mm}
\section{Introduction}

Pedestrian and vehicle trajectory forecasting has seen an increasing interest over the last few years due to the development of a large number of applications in robotics, autonomous driving, video surveillance, and embodied navigation~\cite{Alahi2016SocialLH, Ballan2016KnowledgeTF, Gupta2018SocialGS, Marchetti2020MANTRAMA, Rudenko2019HumanMT}. Accurate prediction of an agent's movements is critical for these applications to ensure safety and improve efficiency. 
Short-term human trajectory forecasting has been extensively explored in the prior literature, and various methods have been proposed to predict the next few seconds of a person's planned intention~\cite{monforte2020and, CuiCWSU23, Camporese2020KnowledgeDF, 10095781}. However, long-term human trajectory forecasting~\cite{Mangalam2020FromGW, loki, Chiara2022GoaldrivenSR}, which involves predicting human motion over a much longer time horizon, remains challenging due to the complexity of human behavior and the uncertainty of future interactions and intentions.
On the other hand, the advent of transformers~\cite{Vaswani2017AttentionIA, Brown2020LanguageMA}, enabled a much broader long-range sequence processing, and in the context of trajectory forecasting, the long-term prediction task gained more attention with transformer-based architectures~\cite{Chiara2022GoaldrivenSR, Yuan2021AgentFormerAT, Franco2023}.
\begin{figure}[t]
    \centering
    \includegraphics[width=0.995\linewidth]{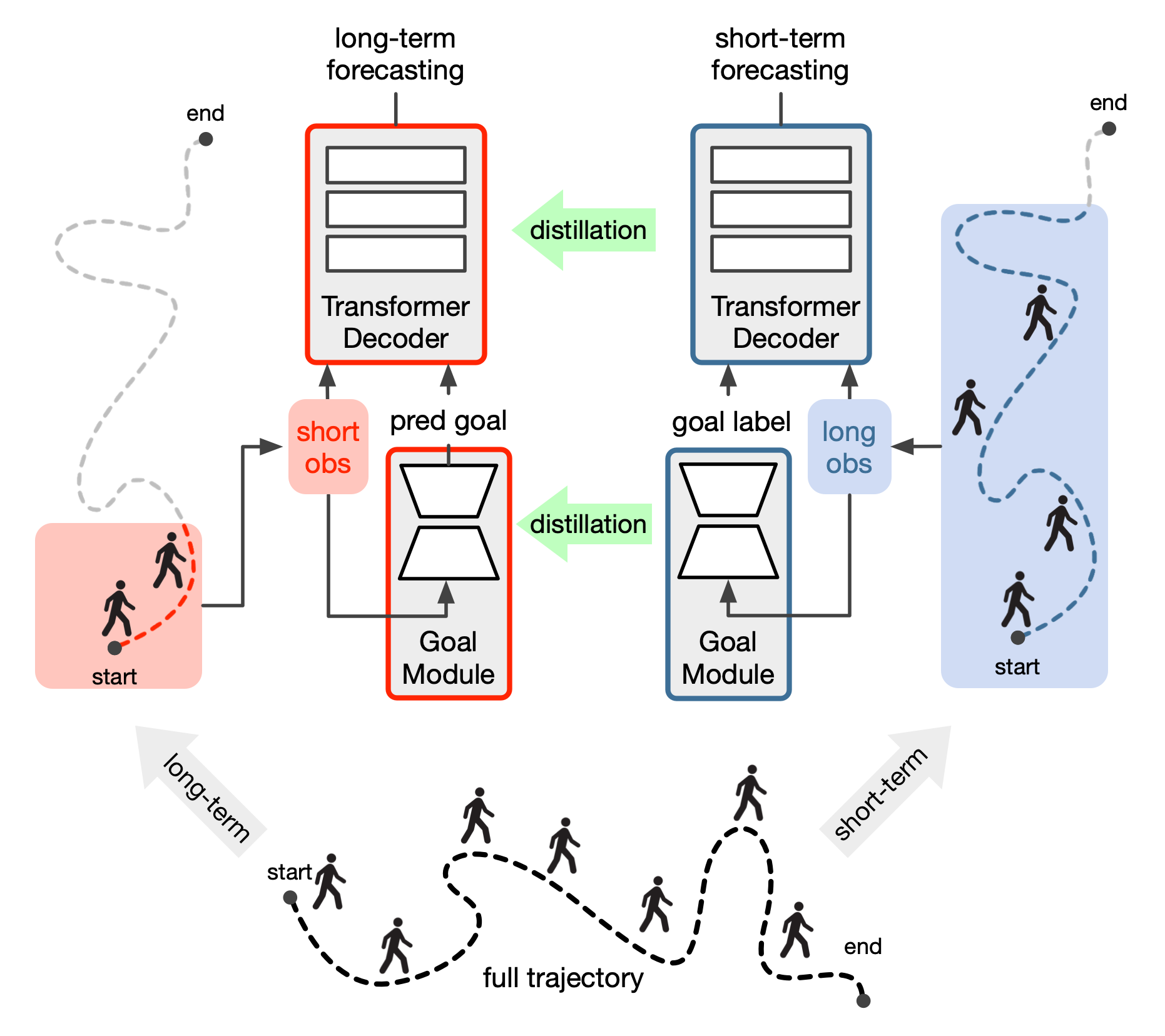}
    \caption{\small{\textbf{Di-Long Framework:} at the bottom, we depict the full trajectory from which the observations and targets of the student and the teacher are extracted. On top we show the components of our framework: the student transformer decoder predicts the long-term trajectory that is distilled from the teacher's prediction (based on longer sequences). The student's decoder is conditioned on the student goal module, distilled from a teacher's goal module.}}
    \label{fig:main_fig}
    \vspace{-8pt}
\end{figure}

In addition to modeling the motion dynamics of individual agents, also contextual information from the environment semantics, from the social interaction with other agents, and from the long-term goals proved their importance in modeling trajectories~\cite{Alahi2016SocialLH, Huang2019STGATMS, Mangalam2020FromGW}. For example,~\cite{Alahi2016SocialLH, Gupta2018SocialGS} proposed methods that model social interactions among agents in the scene with (i) LSTMs that pool agent's information in a neighbor local region of a given location or (ii) employing a GAN for generating socially compliant trajectories.
Furthermore,~\cite{Mangalam2020FromGW} proposed to condition the future predictions with the environment semantic maps that encoded physically compliant trajectories, thus creating a spatially and semantically-aware model. Moreover, the authors proposed to learn long-term and intermediate goals with a specialized U-Net network and to condition future forecasting with them.

In this work, we build upon these ideas and we propose the \emph{Di-Long model} which aims to further improve the performance of long-term trajectory forecasting approaches by reducing the model's future uncertainty with a short-to-long knowledge distillation framework (see Fig.~\ref{fig:main_fig}).
The core idea of knowledge distillation~\cite{Hinton2015DistillingTK} is to transfer knowledge from an expert large model (``teacher'') to a smaller model (``student''), in order to improve the latter with the additional guidance of the expert teacher. 
To this end, we apply knowledge distillation to efficiently train models for long-term forecasting, with a teacher that is an expert in the short-term prediction setting and guides the student on the long-term task during training. 
To the best of our knowledge, this idea has not been explored in the context of long-term trajectory forecasting.
Our paper mostly focuses on human (pedestrian) trajectory prediction, and the Di-Long model shows state-of-the-art performance on two popular datasets, namely the Intersection Drone Dataset (inD) and the Stanford Drone Dataset (SDD), in different forecasting settings.

\begin{figure*}[!ht]
    \centering
    \includegraphics[width=0.9\linewidth]{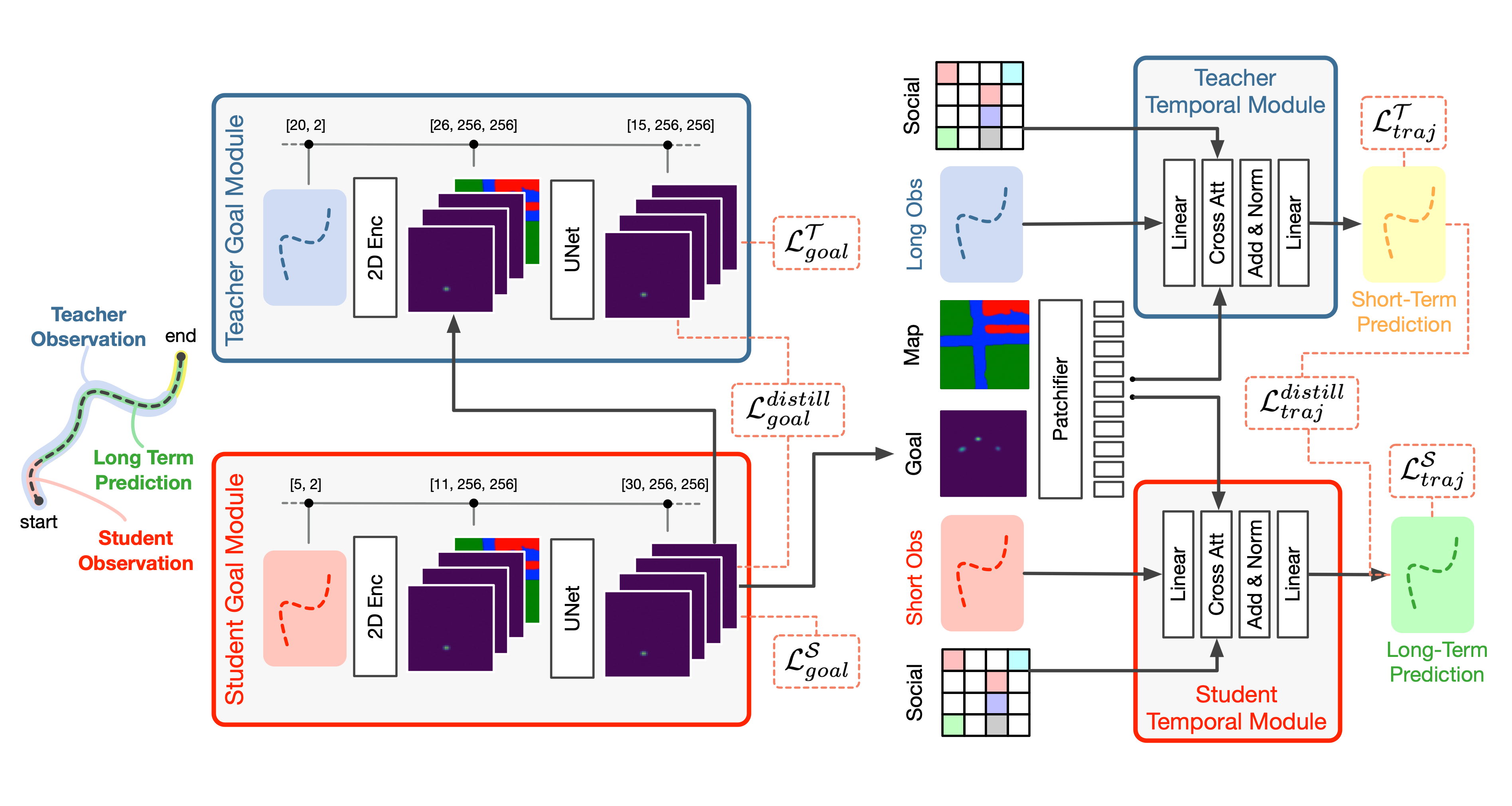}
    \caption{\small{
    \textbf{Detailed overview of the Di-Long Model Components.} Di-Long is composed by a student and the teacher both having a goal module and a temporal module. The student processes short observations and predicts long predictions, the teacher observes long trajectories and predicts short ones. The goal modules processes 2D encoded sequences and semantic maps, producing goal and waypoints heatmaps. The temporal modules, given the observed trajectory, the goals, the semantic maps, and the social information, predict the future trajectory. The distillation is done both in the goal and in the temporal modules. See Sec.~\ref{sec:our_method} for more details.
    }
    }
\label{fig:model}    
\end{figure*}

\section{Related Work}
\label{sec:related}
\textbf{Trajectory Forecasting.}
In recent years, there has been growing interest in predicting the future trajectories of an agent (e.g. pedestrian, vehicle, bicycle, etc.) as accurately as possible, under different settings. The most common is short-term trajectory forecasting, which aims to predict the next few seconds of the agent's path. 
Many methods have been proposed for this task, including recurrent neural networks (RNNs)~\cite{Alahi2016SocialLH, Zhang2019SRLSTMSR}, convolutional neural networks~\cite{Mangalam2020FromGW}, and graph neural networks (GNNs)~\cite{Yu2020SpatioTemporalGT, Salzmann2020TrajectronMG}. 
In addition to modeling the dynamics of individual agent's motion, contextual information such as social interactions has been proven to significantly influence the trajectory prediction performance~\cite{Alahi2016SocialLH, Gupta2018SocialGS, Yuan2021AgentFormerAT}, as well as other contextual features such as physical constraints and scene elements~\cite{Ballan2016KnowledgeTF, Desire2017, Sadeghian2018SoPhieAA}.

Although short-term trajectory forecasting models have achieved remarkable success, long-term forecasting remains challenging. 
The difficulty lies in the complexity of human behavior, which is affected by various factors such as the intent of other agents, randomness in the agents' decisions, the long-term goals of the agents, etc. There are very few approaches, made by the researchers to tackle this, such as estimating multi-modal long-term goals of the agents first, and then randomness is further reduced by sampling intermediate waypoints, conditioned on the final goal~\cite{Mangalam2020FromGW}.
\smallskip

\textbf{Goal-based Trajectory Forecasting.}
Some recent works~\cite{albrecht2021interpretable,zhao2020tnt} employ destination or final goal estimation for improved trajectory prediction. Mangalam \etal~\cite{Mangalam2020ItIN} use a VAE for estimating final goals and employs a prediction network that takes as input the past trajectory and the VAE-predicted goals to predict the future trajectories. 
In a consequent paper~\cite{Mangalam2020FromGW}, they take up a convolutional-based approach where positions are treated as heatmaps and final positions are sampled from 2D probability distribution maps, similar to ours. 
In their recent work, Gilles \etal~\cite{gilles2022gohome} consider directly HD-Maps or lane graphs generated from the HD-Maps, along with the past trajectories to estimate the final goals. However, HD-Maps annotations are expensive and hard to collect.
\smallskip

\textbf{Knowledge Distillation for Trajectory Forecasting.}
In the context of human trajectory forecasting, knowledge distillation has been investigated to improve the model's robustness to incorrect detection and corruption of trajectory data in crowded scenes by distilling knowledge from a teacher with uncorrupted sequence to a student with corrupted observations~\cite{Monti2022HowMO}. In~\cite{wang2023enhancing}, a two-fold knowledge distillation scheme is proposed to transfer more accurate predictability of a teacher network, which has an additional input modality of HD Maps, to a Map-less student network. However, to the best of our knowledge, knowledge distillation has not been explored for long-term trajectory forecasting. In this paper, we propose a novel method for multi-modal long-term trajectory prediction using knowledge distillation.


\vspace{2mm}

\section{Our Method}
\label{sec:our_method}
\subsection{Trajectory Forecasting Problem Formulation}
Given a recorded scene of trajectories $\mathcal{D} = \{\mathcal{I}, \ \mathcal{U} \}$, where $\mathcal{I} \in \mathbb{R}^{3 \times H \times W}$ is the RGB image representing the appearance of the scene, and $\mathcal{U} = \{ \mathbf{u}^i \}_{i=1}^N$ is the collection of all the $N$ agents' trajectories, we are interested in the forecasting of trajectories $t_f$ seconds ahead in the future, from past observations of $t_p$ seconds. More formally, starting from a past observed trajectory $\mathbf{u}^i_p = \{ u^i_k\}_{k=1}^{K_p}$ where $(x^i_k, y^i_k) = u^i_k \in \mathbb{R}^2$ is the 2D $i$-th agent location at the $k$-th step at time $t=k/f_s$ with $f_s$ being the sampling rate of the recording and ${K_p} = t_p\cdot f_s$, we are interested in the estimation of its continuation into the future of $t_p$ seconds defined by $\mathbf{u}^i_f = \{u^i_k\}_{k = K_p+1}^{K_p + K_f}$ where ${K_f} = t_f\cdot f_s$.
Following the long-term setting in~\cite{Mangalam2020FromGW}, we set $t_p = 5$ sec and $t_f = 30 $ sec whereas following the standard practice for the short-term setting~\cite{Gupta2018SocialGS, Huang2019STGATMS, Bertugli2020ACVRNNAC} we set $t_p = 3.2$ sec and $t_f = 4.8$ sec.

\subsection{The Di-Long Model Architecture} 
Pedestrian trajectory forecasting is a challenging task requiring modeling multiple factors influencing human motion. We stress the fact that multi-modality modeling (here instantiated in terms of goal prediction, social influence, and the spatial relation of the trajectory points w.r.t the scene semantic map) are key components of this problem. To this end, we propose a simple
yet powerful architecture that consists of two main components: (i) a \textit{goal estimation module} that predicts the likely final locations of each agent given its previously observed positions, and (ii) a multi-modal, recurrent \textit{temporal backbone} which processes trajectory locations using a cross-attentive mechanism among inputs of different modalities. 
Finally, we have an auxiliary teacher network, which is composed of a similar goal module and a temporal backbone and distills the knowledge about the determinacy of the predicted trajectory to a student network.
\smallskip

\textbf{Goal Estimation Module.}
The goal module $\mathcal{G}$ is a U-Net~\cite{Ronneberger2015UNetCN} that predicts a probability distribution of plausible final positions (i.e. goals) and intermediate waypoints distributions for each input trajectory.
Similar to~\cite{Chiara2022GoaldrivenSR}, observed trajectories $\mathbf{u}_p \in \mathbb{R}^{K_p \times 2}$ are spatially encoded into 2D Gaussian heatmaps by mapping each $u_k$ coordinates into a 2D Gaussian heatmap of shape $H \times W$ centered at $(x_k, y_k)$ with a pre-defined standard deviation $\sigma$, resulting in an tensor $\mathbf{U}_p \in \mathbb{R}^{K_p \times H \times W}$. The same applies to the target trajectory $\mathbf{u}_f \in \mathbb{R}^{K_f \times 2}$ that is encoded into Gaussian heatmaps $\mathbf{U}_f \in \mathbb{R}^{K_f \times H \times W}$.
Other than the 2D encoded trajectory, similar to~\cite{Mangalam2020FromGW} we employ a pre-trained U-Net model $\mathcal{Q}$ to extract the 2D semantic maps of the scene to exploit the physical contextual information useful for the goal module. Specifically, given the scene image $\mathcal{I}$, the pre-trained model provides the segmented one-hot encoded maps $\mathbf{M} \in \mathbb{R}^{C \times H \times W}$ where the set of semantic categories are related to different physical categories of the scene (such as "road", "tree", "terrain", etc.).
We subsequently concatenate the past encoded trajectory $\mathbf{U}_p$ and the estimated semantic maps $\mathbf{M}$ along the channel dimension, and we feed the obtained tensor $\mathbf{x} \in \mathbb{R}^{(K_p + C) \times H \times W}$ to the goal module to produce the goal heatmaps $\mathbf{z} = \mathcal{G}(\mathbf{x}) \in \mathbb{R}^{K_f \times H \times W}$.
In the following, we summarize the computation of the goal module:
\begin{equation}
\label{eq:goal_module_eq}
\begin{split}
    &\mathbf{M} = \mathcal{Q}(\mathcal{I}), \ \ \ \ \ \ \mathbf{U}_p = Gauss2D(\mathbf{u}_p) \\
    &\mathbf{x} = Cat(\mathbf{U}_p, \mathbf{M}), \ \ \ \ \ \ \mathbf{z} = \mathcal{G}(\mathbf{x}) 
\end{split}
\end{equation}
The channel $\mathbf{z}[c] \in \mathbb{R}^{H \times W}$ is a sigmoid-activated probability heatmap related to the specific trajectory forecasting at the time step $c \in [0, K_f - 1]$, and for $c = K_f - 1$ it predicts the final destination (a.k.a. goal) heatmap, whereas for other values of $c=\bar{c}$ it computes an intermediate waypoint distribution in between the last observed point and the goal. 
Following~\cite{Mangalam2020FromGW, Chiara2022GoaldrivenSR}, to reduce the future uncertainty, we sample multiple goals and waypoints from the corresponding heatmap channels of $\mathbf{z}$, and we recompute the Gaussian encoded maps that are used by the temporal backbone module. In our investigation, similar to~\cite{Chiara2022GoaldrivenSR} we make use of the goal $\mathbf{z}[K_f-1]$ and a single waypoint $\mathbf{z}[\bar{c}]$ at $\bar{c} = K_f // 2$, and we subsequently channel-wise concatenate them into $\mathbf{\hat{z}} \in \mathbb{R}^{2 \times H \times W}$.

\begin{table*}[ht]
\centering
\resizebox{0.76\linewidth}{!}{%
\begin{tabular}{lcccccccc}
    \toprule
    \textbf{Method} & \textbf{Social} & \textbf{Semantic Map} & \textbf{Goal} & \textbf{Distill.} & \textbf{ADE} & \textbf{FDE} & \textbf{KDE-NLL} & \textbf{Miss Rate}\\
    \cmidrule{1-9}
    Social-GAN~\cite{Gupta2018SocialGS} & \color{green!60!black}\ding{51} & \color{red}\ding{55} & \color{red}\ding{55} & \color{red}\ding{55} & 38.57 & 84.61 & - & -\\
    PECNet~\cite{Mangalam2020ItIN} & \color{green!60!black}\ding{51} & \color{red}\ding{55} & \color{green!60!black}\ding{51} & \color{red}\ding{55} & 20.25 & 32.95 & - & -\\
    R-PECNet~\cite{Mangalam2020ItIN, Mangalam2020FromGW} & \color{green!60!black}\ding{51} & \color{red}\ding{55} & \color{green!60!black}\ding{51} & \color{red}\ding{55} & 341.80 & 1702.64 & - & -\\
    Social-STGCNN$^\dagger$~\cite{mohamed2020social}  & \color{green!60!black}\ding{51} & \color{red}\ding{55} & \color{red}\ding{55} & \color{red}\ding{55} & 22.48\scriptsize{$\pm$0.14} & 33.61\scriptsize{$\pm$0.29} & 13.67\scriptsize{$\pm$0.01} & 0.58\scriptsize{$\pm$0.01}\\
    Agentformer$^\dagger$~\cite{Yuan2021AgentFormerAT}  & \color{green!60!black}\ding{51} & \color{red}\ding{55} & \color{red}\ding{55} & \color{red}\ding{55} & 36.26\scriptsize{$\pm$0.15} & 46.36\scriptsize{$\pm$0.34} & 19.53\scriptsize{$\pm$0.01} & 0.63\scriptsize{$\pm$0.01}\\    
    Trajectron++$^\dagger$~\cite{Salzmann2020TrajectronMG} & \color{green!60!black}\ding{51} & \color{green!60!black}\ding{51} & \color{red}\ding{55} & \color{red}\ding{55} & 24.16\scriptsize{$\pm$0.13} & 36.86\scriptsize{$\pm$0.33} & 15.45\scriptsize{$\pm$0.02} & 0.65\scriptsize{$\pm$0.01}\\
    Goal-SAR$^\dagger$~\cite{Chiara2022GoaldrivenSR} & \color{red}\ding{55} & \color{red}\ding{55} & \color{green!60!black}\ding{51} & \color{red}\ding{55} & 18.53\scriptsize{$\pm$0.10} & 26.59\scriptsize{$\pm$0.37} & 9.80\scriptsize{$\pm$0.01} & 0.34\scriptsize{$\pm$0.01}\\
    Y-Net$^*$~\cite{Mangalam2020FromGW} & \color{red}\ding{55} & \color{green!60!black}\ding{51} & \color{green!60!black}\ding{51} & \color{red}\ding{55} & \underline{15.19\scriptsize{$\pm$0.13}} & \underline{23.17\scriptsize{$\pm$0.31}} & 10.17\scriptsize{$\pm$0.01} & \underline{0.32\scriptsize{$\pm$0.02}}\\
    \cmidrule[0.01mm]{1-9}
    Di-Long (Ours) & \color{green!60!black}\ding{51} & \color{green!60!black}\ding{51} & \color{green!60!black}\ding{51} & \color{red}\ding{55} & 17.79\scriptsize{$\pm$0.13} & 23.33\scriptsize{$\pm$0.42} & \underline{9.73\scriptsize{$\pm$0.02}} & 0.36\scriptsize{$\pm$0.01}\\
    \textbf{Di-Long (Ours)} & \color{green!60!black}\ding{51} & \color{green!60!black}\ding{51} & \color{green!60!black}\ding{51} & \color{green!60!black}\ding{51} & \textbf{14.89\scriptsize{$\pm$0.15}} & \textbf{19.85\scriptsize{$\pm$0.28}} & \textbf{9.52\scriptsize{$\pm$0.01}} & \textbf{0.24\scriptsize{$\pm$0.01}}\\
    \bottomrule
\end{tabular}
}
\caption{\small{\textbf{Intersection Drone Dataset (inD) Long-term Results.} Results are reported in terms of the best ADE and FDE among $20$ predicted generated samples, in meter (lower is better). The KDE-NLL is calculated by averaging over $20$ predictions. Miss Rate is calculated over 20 predictions with a center distance threshold of 20 meter. Bold and underlined numbers indicate the best and second-best from previous works. \textbf{Social}, \textbf{Semantic Map}, \textbf{Goal} indicate additional input information (other than temporal) used by the models. \textbf{Distill.} is not a modality but an enhancement to the model. \emph{$\dagger$ means that we re-trained the model as the authors do not provide the results in the long-term setting; $*$ means that we re-trained the model in order to align the data pre-processing w.r.t. all the other previous works.}}}
\label{tab:long_term_ind}
\end{table*}

\begin{table*}[ht]
\centering
\resizebox{0.76\linewidth}{!}{%
\begin{tabular}{lcccccccc}
    \toprule
    \textbf{Method} & \textbf{Social} & \textbf{Semantic Map} & \textbf{Goal} & \textbf{Distill.} & \textbf{ADE} & \textbf{FDE} & \textbf{KDE-NLL} & \textbf{Miss Rate}\\
    \cmidrule{1-9}
    Social-GAN~\cite{Gupta2018SocialGS} & \color{green!60!black}\ding{51} & \color{red}\ding{55} & \color{red}\ding{55} & \color{red}\ding{55} & 155.32 & 307.88 & - & -\\
    PECNet~\cite{Mangalam2020ItIN} & \color{green!60!black}\ding{51} & \color{red}\ding{55} & \color{green!60!black}\ding{51} & \color{red}\ding{55} & 72.22 & 118.13 & - & -\\
    R-PECNet~\cite{Mangalam2020ItIN, Mangalam2020FromGW} & \color{green!60!black}\ding{51} & \color{red}\ding{55} & \color{green!60!black}\ding{51} & \color{red}\ding{55} &  261.27 & 750.42 & - & -\\
    Social-STGCNN$^\dagger$~\cite{mohamed2020social}  & \color{green!60!black}\ding{51} & \color{red}\ding{55} & \color{red}\ding{55} & \color{red}\ding{55} & 69.89\scriptsize{$\pm$0.09} & 112.73\scriptsize{$\pm$0.14} & 13.24\scriptsize{$\pm$0.01} & 0.37\scriptsize{$\pm$0.01}\\
    Agentformer$^\dagger$~\cite{Yuan2021AgentFormerAT}  & \color{green!60!black}\ding{51} & \color{red}\ding{55} & \color{red}\ding{55} & \color{red}\ding{55} & 66.43\scriptsize{$\pm$0.10} & 102.60\scriptsize{$\pm$0.16} & 12.13\scriptsize{$\pm$0.01} & 0.35\scriptsize{$\pm$0.01}\\    
    Trajectron++$^\dagger$~\cite{Salzmann2020TrajectronMG} & \color{green!60!black}\ding{51} & \color{green!60!black}\ding{51} & \color{red}\ding{55} & \color{red}\ding{55} & 71.51\scriptsize{$\pm$0.13} & 130.41\scriptsize{$\pm$0.13} & 14.68\scriptsize{$\pm$0.01} & 0.55\scriptsize{$\pm$0.01}\\
    Goal-SAR$^\dagger$~\cite{Chiara2022GoaldrivenSR} & \color{red}\ding{55} & \color{red}\ding{55} & \color{green!60!black}\ding{51} & \color{red}\ding{55} & 53.08\scriptsize{$\pm$0.04} & 79.26\scriptsize{$\pm$0.10} & 11.53\scriptsize{$\pm$0.01} & 0.31\scriptsize{$\pm$0.01}\\
    Y-Net$^*$~\cite{Mangalam2020FromGW} & \color{red}\ding{55} & \color{green!60!black}\ding{51} & \color{green!60!black}\ding{51} & \color{red}\ding{55} & \textbf{46.00\scriptsize{$\pm$0.23}} & \underline{77.45\scriptsize{$\pm$0.85}} & 11.23\scriptsize{$\pm$0.01} & \textbf{0.26\scriptsize{$\pm$0.01}}\\
    \cmidrule{1-9}
    Di-Long (Ours) & \color{green!60!black}\ding{51} & \color{green!60!black}\ding{51} & \color{green!60!black}\ding{51} & \color{red}\ding{55} & 48.92\scriptsize{$\pm$0.09} & 76.31\scriptsize{$\pm$0.18} & \underline{10.93\scriptsize{$\pm$0.01}} & 0.34\scriptsize{$\pm$0.01}\\
    \textbf{Di-Long (Ours)} & \color{green!60!black}\ding{51} & \color{green!60!black}\ding{51} & \color{green!60!black}\ding{51} & \color{green!60!black}\ding{51} & \underline{48.21\scriptsize{$\pm$0.09}} & \textbf{72.41\scriptsize{$\pm$0.21}} & \textbf{10.85\scriptsize{$\pm$0.01}} & \underline{0.28\scriptsize{$\pm$0.01}}\\
    \bottomrule
\end{tabular}
}
\caption{\small{\textbf{Stanford Drone Dataset (SDD) Long-term Results.} Results are reported as the best ADE and FDE among $20$ predicted samples in pixels (lower is better). The KDE-NLL is calculated by averaging over $20$ predictions. Miss Rate is calculated over 20 predictions with a center distance threshold of 80 pixels. \emph{$\dagger$ and $*$ should be interpreted as previously reported in Table~\ref{tab:long_term_ind}.}}}
\label{tab:long_term_Robicquet2016LearningSE}
\vspace{-4mm}
\end{table*}

\textbf{Temporal Module.}
Similar to~\cite{Chiara2022GoaldrivenSR}, our temporal module $T$ is an autoregressive transformer that processes the past observed coordinates $\mathbf{u}_p \in \mathbb{R}^{K_p \times 2}$ and predicts the future trajectory $\mathbf{v} \in \mathbb{R}^{K_f \times 2}$ related to the ground truth $\mathbf{u}_f \in \mathbb{R}^{K_f \times 2}$. However, in contrast to~\cite{Chiara2022GoaldrivenSR}, we employ a transformer decoder-only (GPT~\cite{Brown2020LanguageMA} style) that uses cross-attention modules to be conditioned by several input modalities such as goal and waypoints estimated maps, scene semantic maps, and social information from other agents in the scene. 
More specifically, the predicted semantic map of the scene $\mathcal{Q}(\mathcal{I}) = \mathbf{M} \in \mathbb{R}^{C \times H \times W}$ is channel-wise concatenated to the estimated sampled maps $\mathbf{\hat{z}}$ (containing the goal $\mathbf{z}[K_f]$ and the waypoints $\mathbf{z}[\bar{c}]$) and together are patchified, linearly projected, and fed to the autoregressive transformer decoder.  
Inspired by~\cite{Yuan2021AgentFormerAT}, in our temporal module, we employ an agent-aware cross-attention between the past observed trajectory and the concatenated multi-modal information (semantic map, goals, and waypoints). This attention mechanism preserves the notion of agent identity and learns the intra and inter-dependencies among all the agents~\cite{Yuan2021AgentFormerAT}. 
In the following, we summarize the temporal module forward computation:
\begin{equation}
\label{eq:temp_module_eq}
\begin{gathered}
    \mathbf{\hat{u}}_\mathcal{G} = Proj(Patch(Cat(\mathbf{\hat{z}}, \mathbf{M}))) \\
    \mathbf{v} = T(\mathbf{u}_p, \mathbf{\hat{u}}_\mathcal{G})
\end{gathered}
\end{equation}
where $Proj(\cdot)$ is a linear projection layer and $Patch(\cdot)$ is the standard patchification operation of the vision transformer~\cite{dosovitskiy2020image} that tokenizes a 2D tensor into tokens.

\subsection{Short-to-Long Term Knowledge Distillation} 
In our framework, we create two instances of the same model for the teacher $\mathcal{T}$ and the student $\mathcal{S}$ where both have the previously described goal and temporal modules. However, we let the two networks process different observations with different duration (longer for the teacher and shorter for the student), and we distill the information of the teacher to guide the student's predictions both for the goal and temporal modules. 
More specifically, the teacher $\mathcal{T}$ observes trajectories of length $t_p + t_a$ seconds with $0 \leq t_a < t_f$ while the student $\mathcal{S}$ processes trajectories of duration $t_p$. We define $t_a$ as the anchor time, and as $t_a$ increases, the teacher solves a less uncertain task since its observation grows and the length of its prediction $t_f - t_a$ becomes smaller and more deterministic. In the following, we consider $t_a = K_f//2$ as investigated in Sec.~\ref{sec:ablation}. $\mathcal{T}$ is only used at training time, and during inference, only $\mathcal{S}$ is considered.

\textbf{Goal Module Distillation.} 
Trajectory forecasting is heavily dependent on the predicted goal~\cite{Mangalam2020FromGW, Chiara2022GoaldrivenSR} as a better goal estimation leads to a better-predicted trajectory closer to the ground truth. For this reason, we focused on the improvement of our goal module $\mathcal{G}$ by following a knowledge distillation strategy. 
Following Eq.~\ref{eq:goal_module_eq}, the student goal module processes $\mathbf{x}^{\mathcal{S}} \in \mathbb{R}^{(K_p + C) \times H \times W}$ producing $\mathbf{z}^\mathcal{S} \in \mathbb{R}^{K_f \times H \times W}$, whereas the teacher processes $\mathbf{x}^{\mathcal{T}} \in \mathbb{R}^{ (K_p + K_a + C) \times H \times W}$ producing $\mathbf{z}^\mathcal{T} \in \mathbb{R}^{(K_f - K_a) \times H \times W}$, with $K_a = t_a \cdot f_s$. 
The teacher and student goal modules are then trained by optimizing their corresponding losses as defined in the following:
\begin{equation}
    \begin{gathered}
    \mathcal{L}_{goal}^{\mathcal{S}} = \frac{1}{K_f} \sum_{k=0}^{K_f-1} BCE \left( \mathbf{z}^{\mathcal{S}}[k], \mathbf{U}_f[k] \right) \\
    \mathcal{L}_{goal}^{\mathcal{T}} = \frac{1}{K_f - K_a} \sum_{k=0}^{\mathclap{K_f - K_a - 1}} \ BCE \left( \mathbf{z}^{\mathcal{T}}[k], \mathbf{U}_f[K_a + k] \right)
    \end{gathered}
\end{equation}
where $BCE(p, t)$ is the binary cross-entropy loss between a generic prediction $p$ and target $t$.

In our investigation, the teacher and the student are trained from scratch together in an online fashion. Interestingly, during training, we found it beneficial to distill information from the teacher in an \textit{implicit} manner without using the usual explicit distillation loss~\cite{Hinton2015DistillingTK, Monti2022HowMO}. Specifically, we let the teacher see the standard observation $\mathbf{x}^{\mathcal{T}}$ as well as an augmented version of it $\mathbf{x}^{\mathcal{T}\mathcal{S}}$ by replacing the last portion of the observation (from $K_p$ up to $K_p + K_a$) with the time-related student's prediction counterpart. More specifically:
\begin{equation}
\begin{gathered}
    \mathbf{x}^{\mathcal{T}\mathcal{S}}[0:K_p] = \mathbf{x}^{\mathcal{T}}[0:K_p], \\
    \mathbf{x}^{\mathcal{T}\mathcal{S}}[K_p:K_p+K_a] = \mathbf{z}^{\mathcal{S}}[0:K_a].
\end{gathered}
\end{equation}
In this way, the teacher provides a stronger regularization for the student with the guidance of a teacher who is \textit{informed} of the current student's prediction state, and the student is updated with the gradients that pass through the teacher's predictions.
Given the two views $\mathbf{x}^{\mathcal{T}}$ and $\mathbf{x}^{\mathcal{T}\mathcal{S}}$  the teacher goal module produces respectively $\mathbf{z}^{\mathcal{T}}$ and $\mathbf{z}^{\mathcal{T}\mathcal{S}}$, and we define the distillation loss as:
\begin{equation}
   \mathcal{L}^{distill}_{goal} = \frac{1}{K_f - K_a} \sum_{k=0}^{\mathclap{K_f - K_a - 1}} \ BCE \left( \mathbf{z}^{\mathcal{T}\mathcal{S}}[k], \mathbf{U}_f[k] \right)
\end{equation}
Finally, the total loss for training the goal modules in our knowledge distillation setting results as in the following:
\begin{equation}    
    \mathcal{L}_{goal} = \mathcal{L}^{\mathcal{S}}_{goal} + \mathcal{L}^{\mathcal{T}}_{goal} + \mathcal{L}^{distill}_{goal}.
\end{equation}

\begin{table}[!t]
\centering
\resizebox{0.48\textwidth}{!}{%
\begin{tabular}{lccccc}
    \toprule
    \textbf{Method} & \textbf{Distill.} & \textbf{ADE} & \textbf{FDE} & \textbf{KDE-NLL} & \textbf{MR}\\
    \cmidrule{1-6}
    Social-GAN~\cite{Gupta2018SocialGS} & \color{red}\ding{55} & 27.23 & 41.44 & - & -\\
    CF-VAE~\cite{Bhattacharyya2019ConditionalFV} & \color{red}\ding{55} & 12.60 & 22.30 & - & -\\
    P2T~\cite{Deo2020TrajectoryFI} & \color{red}\ding{55} & 12.58 & 22.07 & - & -\\
    SimAug~\cite{Liang2020SimAugLR} & \color{red}\ding{55} & 10.27 & 19.71 & - & -\\
    PECNet~\cite{Mangalam2020ItIN} & \color{red}\ding{55} & 9.96 & 15.88 & - & -\\
    Social-STGCNN$^\dagger$~\cite{mohamed2020social}  & \color{red}\ding{55} & 18.88\scriptsize{$\pm$0.01} & 33.14\scriptsize{$\pm$0.01} & 9.32\scriptsize{$\pm$0.01} & 0.56\scriptsize{$\pm$0.01}\\
    Agentformer$^\dagger$~\cite{Yuan2021AgentFormerAT} & \color{red}\ding{55} & 10.25\scriptsize{$\pm$0.01} & 16.51\scriptsize{$\pm$0.03} & 9.56\scriptsize{$\pm$0.01} & 0.37\scriptsize{$\pm$0.01}\\
    Trajectron++$^\dagger$~\cite{Salzmann2020TrajectronMG} & \color{red}\ding{55} & 10.29\scriptsize{$\pm$0.01} & 15.98\scriptsize{$\pm$0.02} & 8.10\scriptsize{$\pm$0.01} & 0.32\scriptsize{$\pm$0.01}\\
    Goal-SAR$^*$~\cite{Chiara2022GoaldrivenSR} & \color{red}\ding{55} & \underline{7.98\scriptsize{$\pm$0.01}} & 12.21\scriptsize{$\pm$0.03} & \underline{7.93\scriptsize{$\pm$0.01}} & \underline{0.23\scriptsize{$\pm$0.01}}\\
    Y-Net$^*$~\cite{Mangalam2020FromGW} & \color{red}\ding{55} & 8.25\scriptsize{$\pm$0.01} & \underline{12.10\scriptsize{$\pm$0.02}} & 8.22\scriptsize{$\pm$0.01} & 0.25\scriptsize{$\pm$0.01}\\
    \cmidrule{1-6}
    Di-Long (Ours) & \color{red}\ding{55} & \textbf{7.43\scriptsize{$\pm$0.01}} & 12.13\scriptsize{$\pm$0.01} & 8.05\scriptsize{$\pm$0.01} & \textbf{0.20\scriptsize{$\pm$0.01}}\\
    \textbf{Di-Long (Ours)} & \color{green!60!black}\ding{51} & \textbf{7.43\scriptsize{$\pm$0.01}} & \textbf{12.07\scriptsize{$\pm$0.01}} & \textbf{7.85\scriptsize{$\pm$0.01}} & \textbf{0.20\scriptsize{$\pm$0.01}} \\
    \bottomrule
\end{tabular}
}
\caption{\small{\textbf{Short-term results on the SDD.} Miss Rate (MR) is calculated over 20 predictions with a center distance threshold of 14 pixels. $*$ means that we re-trained the models in order to align the data pre-processing w.r.t. all the other previous works. }}
\label{tab:short_term_sdd}
\vspace{-2mm}
\end{table}

\begin{table}[!t]
\centering
\resizebox{0.48\textwidth}{!}{%
\begin{tabular}{lccccc}
    \toprule
    \textbf{Method} & \textbf{Distill.} & \textbf{ADE} & \textbf{FDE} & \textbf{KDE-NLL} & \textbf{MR}\\
    \cmidrule{1-6}
    Social-GAN~\cite{Gupta2018SocialGS} & \color{red}\ding{55} & 0.48 & 0.99 & - & - \\
    ST-GAT~\cite{Huang2019STGATMS} & \color{red}\ding{55} & 0.48 & 1.00 & - & - \\
    AC-VRNN~\cite{Bertugli2020ACVRNNAC} & \color{red}\ding{55} & 0.42 & 0.80 & - & -\\
    Social-STGCNN$^\dagger$~\cite{mohamed2020social}  & \color{red}\ding{55} & 0.59\scriptsize{$\pm$0.01} & 0.96\scriptsize{$\pm$0.01} & 7.97\scriptsize{$\pm$0.01} & 0.61\scriptsize{$\pm$0.01}\\
    Agentformer$^\dagger$~\cite{Yuan2021AgentFormerAT} & \color{red}\ding{55} & 0.57\scriptsize{$\pm$0.01} & 0.87\scriptsize{$\pm$0.01} & 6.86\scriptsize{$\pm$0.01} & 0.53\scriptsize{$\pm$0.01}\\
    Trajectron++$^\dagger$~\cite{Salzmann2020TrajectronMG} & \color{red}\ding{55} & 0.62\scriptsize{$\pm$0.01} & 0.98\scriptsize{$\pm$0.01} & 8.13\scriptsize{$\pm$0.01} & 0.64\scriptsize{$\pm$0.01}\\
    Goar-SAR$^*$~\cite{Chiara2022GoaldrivenSR} & \color{red}\ding{55} & 0.44\scriptsize{$\pm$0.01} & 0.70\scriptsize{$\pm$0.01} & \underline{5.47\scriptsize{$\pm$0.01}} & 0.49\scriptsize{$\pm$0.01}\\
    Y-Net$^*$~\cite{Mangalam2020FromGW} & \color{red}\ding{55} & 0.55\scriptsize{$\pm$0.01} & 0.93\scriptsize{$\pm$0.01} & 7.20\scriptsize{$\pm$0.01} & 0.60\scriptsize{$\pm$0.01}\\
    \cmidrule{1-6}
    Di-Long (Ours) & \color{red}\ding{55} & \underline{0.39\scriptsize{$\pm$0.01}} & \underline{0.61\scriptsize{$\pm$0.01}} & 5.95\scriptsize{$\pm$0.01} & \underline{0.27\scriptsize{$\pm$0.01}}\\
    \textbf{Di-Long (Ours)} & \color{green!60!black}\ding{51} & \textbf{0.37\scriptsize{$\pm$0.01}} & \textbf{0.59\scriptsize{$\pm$0.01}} & \textbf{5.32\scriptsize{$\pm$0.01}} & \textbf{0.25\scriptsize{$\pm$0.01}}\\
    \bottomrule
\end{tabular}
}
\caption{\small{\textbf{Short-term results on the inD.} Miss Rate (MR) is calculated over 20 predictions with a center distance threshold of 5 meter. $*$ means that we re-trained the models in order to align the data pre-processing w.r.t. all the other previous works.
}}
\label{tab:short_term_ind}
\vspace{-5mm}
\end{table}

\vspace{3mm}
\noindent
\textbf{Temporal Module Distillation.} 
Similarly to the goal modules, we employ a knowledge distillation strategy from the temporal module of the teacher $\mathcal{T}$ to the temporal module of the student $\mathcal{S}$ in order to provide the student useful guidance from the teacher that processes a longer observation and solves a less uncertain and easy task. Following Eq.~\ref{eq:temp_module_eq}, the student temporal module takes as input $\mathbf{u}_p^{\mathcal{S}} \in \mathbb{R}^{K_p \times 2}$ as well as $\mathbf{\hat{z}}$ and $\mathbf{M}$ and produces $\mathbf{v}^{\mathcal{S}} \in \mathbb{R}^{K_p \times 2}$. Similarly, the teacher processed $\mathbf{u}_p^{\mathcal{T}} \in \mathbb{R}^{(K_p + K_a) \times 2}$, $\mathbf{\hat{z}}$ and $\mathbf{M}$, and produces $\mathbf{v}^{\mathcal{T}} \in \mathbb{R}^{(K_f - K_a) \times 2}$. In our investigation, for the temporal module, we use a more standard knowledge distillation mechanism by employing an explicit loss term on the teacher and student predictions.
In the following, we summarize all the losses we designed for training the teacher and student temporal modules:
\begin{equation}
\begin{gathered}
    \mathcal{L}^{\mathcal{S}}_{traj} = \frac{1}{K_f} \sum_{k=0}^{K_f - 1} \left\Vert \mathbf{v}^{\mathcal{S}}[k] - \mathbf{u}_f[k] \right\Vert_2^2 \\
    \mathcal{L}^{\mathcal{T}}_{traj} = \frac{1}{K_f - K_a} \sum_{k=0}^{\mathclap{K_f - K_a - 1}} \left\Vert \mathbf{v}^{\mathcal{T}}[k] - \mathbf{u}_f[K_a + k] \right\Vert_2^2 \\
    \mathcal{L}^{distill}_{traj} = \frac{1}{K_f - K_a} \sum_{k=0}^{\mathclap{K_f - K_a - 1}} \left\Vert \mathbf{v}^{\mathcal{S}}[K_a + k] - \mathbf{v}^{T}[k] \right\Vert_2^2.
\end{gathered}
\end{equation}
Subsequently, we define the total loss related to the temporal modules as:
\begin{equation}
    \mathcal{L}_{traj} = \mathcal{L}^{\mathcal{S}}_{traj} + \mathcal{L}^{\mathcal{T}}_{traj} + \mathcal{L}^{distill}_{traj}.
\end{equation}

Finally, we combine the goal and the temporal losses, as well as their distillation loss terms together, and the total loss results:
\begin{equation}
    {\mathcal{L}} = {\mathcal{L}}_{goal} + \lambda {\mathcal{L}}_{traj}
    \label{eq:loss}
\end{equation}
where $\lambda \in \mathbb{R}$ is a mixing hyper-parameter that balances the importance of the goal estimation and the temporal module prediction.


\section{Experiments}

\subsection{Experimental Setting} 
\noindent
\textbf{Long and Short-Term Setting.} 
We follow the long-term forecasting setting of previous works~\cite{Mangalam2020FromGW, Chiara2022GoaldrivenSR} where the observation is $t_p = 5$ sec and the future target trajectory has duration is $t_f = 30$ sec, whereas for the short-time setting we follow the well-established protocol in which $t_p = 3.2$ sec and $t_f = 4.8$ sec (e.g.~\cite{Alahi2016SocialLH, Gupta2018SocialGS}).

\vspace{2mm}
\noindent
\textbf{Datasets.} 
We used two popular datasets for trajectory prediction: Intersection Drone Dataset (inD)~\cite{Bock2019TheID} and Stanford Drone Dataset (SDD)~\cite{Robicquet2016LearningSE}. 

The inD dataset contains $10$ hours of recordings of $4$ different intersections in an urban environment, where pedestrians interact with cars to reach their destinations. The dataset has sample rate $25$ fps and we follow the same pre-processing of~\cite{Yu2020SpatioTemporalGT} to prepare the dataset for the long and short-term settings, such as resampling the sequences at $f_s$ fps ($f_s=1$ fps in the long-term scenario, and $f_s = 2.5$ for the short-term ones), consider only pedestrian tracks, remove short sequences, and use sliding window without overlap to split long trajectories. We follow the standard dataset splits and evaluation setting of~\cite{Mangalam2020FromGW}.

SDD contains $11,000$ unique pedestrian tracks across $20$ top-down scenes from the Stanford campus in a bird's eye view captured with a drone. We follow the dataset splits and pre-processing from~\cite{Mangalam2020ItIN}, where the initial tracks at $25$ fps are resampled at $f_s = 1$ fps for the long-term setting, and at $f_s = 2.5$ fps for the short-term setting.
For both datasets we obtained sequences of length $K_p = 5$, $K_f = 30$ for the long-term and $K_p = 8$, $K_f = 12$ for the short-term scenarios.

\vspace{2mm}
\noindent
\textbf{Metrics.} 
For the quantitative evaluation, we consider two standard error metrics, namely the Average Displacement Error (ADE) and the Final Displacement Error (FDE). The ADE is calculated by measuring the average $\ell_2$ distance between the predicted and ground truth trajectories, while for the FDE, only the final positions are considered. Following previous works~\cite{Mangalam2020FromGW, Chiara2022GoaldrivenSR} that provided the analysis in a stochastic setting, we reported the best-of-$K$ ADE and FDE over $K=20$ generated predictions for each input trajectory. 

Additionally, to measure the correctness of the output predicted distribution, we consider the Kernel Density Estimate Negative Log Likelihood (KDE-NLL) metric introduced by~\cite{Ivanovic2018TheTP}, where the output pdf is first estimated at each prediction step, and subsequently used to compute the average log-likelihood. A trajectory forecast is considered as a miss if the FDE between the ground truth and the predicted trajectory is above a center distance threshold. We calculate Miss Rate~\cite{xu2024towards} over 20 predictions and empirically choose the center distance threshold values for different settings for different datasets. For inD long-term, the threshold is 20 meters; for inD short-term, it is 5 meters. On the other hand, for SDD long-term, the threshold is 80 pixels; for SDD short-term, it is 14 pixels. As for the short-term setting, the predicted trajectories deviate less from the ground truth compared to the long-term setting, we choose a lower threshold value for calculating Miss Rate in the short-term than long-term setting. Compared to inD, SDD contains more complex scenarios and hence higher FDE values are observed in SDD. That is why we define tighter thresholds for trajectory forecasting on inD, compared to SDD.    

\subsection{Experimental Results} 
\noindent
\textbf{Intersection Drone Dataset.} 
In Table~\ref{tab:long_term_ind}, we reported our results on the Intersection Drone Dataset in the long-term setting. 
In this scenario, our Di-Long model outperforms Goal-SAR and Y-Net by a considerable margin in terms of ADE ($-0.3$ w.r.t. Y-Net), FDE ($-3.32$ w.r.t. Y-Net), KDE-NLL ($-0.28$ w.r.t. Goal-SAR) and Miss Rate ($-0.3$ w.r.t. Y-Net). Most importantly, we show that our distillation framework is beneficial for our Di-Long model as, without enabling it, our model has higher ADE, and comparable FDE, KDE-NLL, and Miss Rate w.r.t. the second best previous models (underlined results in the table). In Table~\ref{tab:short_term_ind} we reported the results for the same dataset for the short-term setting. Also in this case, our model outperforms previous works in all the metrics, however, the distillation improvement is smaller, probably due to the short-term setting where the future uncertainty that can be reduced by distillation by a teacher is less prominent.

\begin{figure*}[ht]
    \centering
    \includegraphics[width=0.9\textwidth]{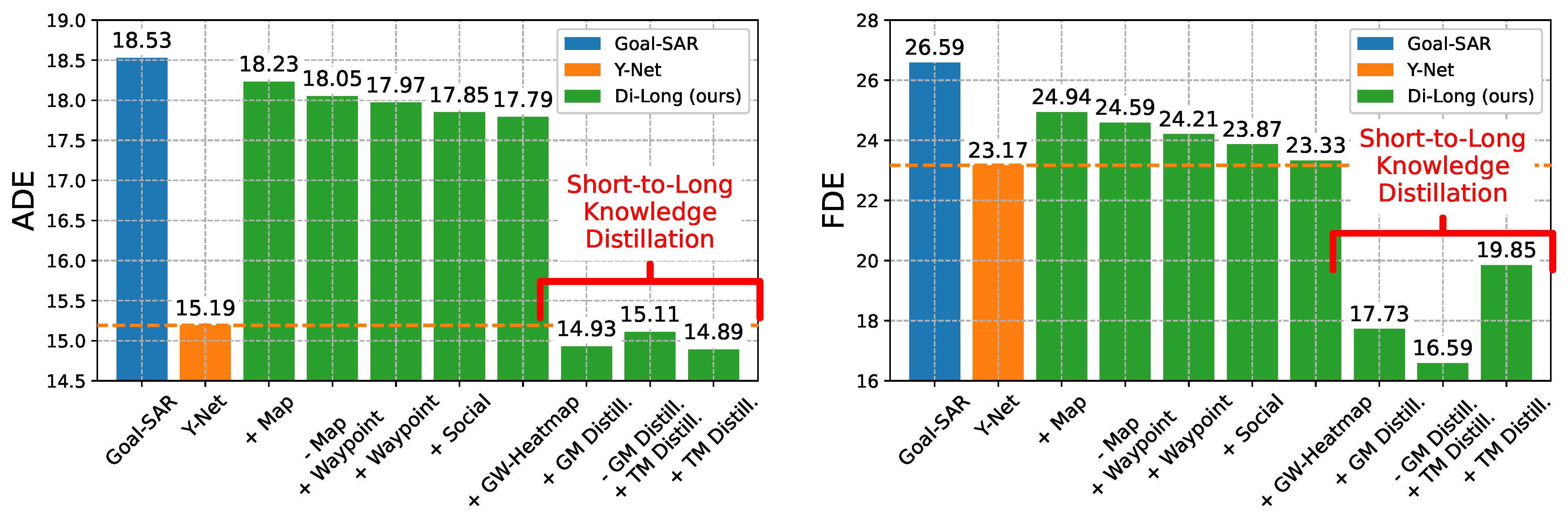}
    \vspace{-3mm}
    \caption{\small{\textbf{Ablation Study on the Di-Long Components.} 
    In \textbf{GW-Heatmap}, a 2D Gaussian heatmap of the goal/waypoint (GW) is appended to the semantic map, patchified, projected, and passed as the control input of the transformer decoder. \textbf{GM Distill} corresponds to goal module distillation, while \textbf{TM Distill} corresponds to temporal module distillation, respectively. 
    \vspace{-3mm}}}
    \label{fig:ablation_dilong}
\end{figure*}

\vspace{2mm}
\noindent
\textbf{Stanford Drone Dataset.} 
In Table~\ref{tab:long_term_Robicquet2016LearningSE}, we reported our results on the Stanford Drone Dataset in long-term settings. Similar to the inD dataset, our model shows impressive performances in terms of FDE ($-5.04$), which suggests that our goal module combined with our distillation strategy (where the distillation gain is $-3.9$ FDE) is crucial to improve the final prediction. We also improve the KDE-NLL measure ($-0.38$) w.r.t. the second best model (results underlined in the table), however, we slightly degrade the ADE and Miss Rate performances w.r.t. Y-Net.
In Table~\ref{tab:short_term_sdd}, we show our results on SDD in the short-term settings. 
Similarly to the inD short-term results, the distillation gain is smaller compared to the long-term setting, however, we confirm to improve the results in terms of ADE, FDE and KDE-NLL w.r.t. previous works.

\subsection{Ablation Studies}
\label{sec:ablation}
\noindent
\textbf{Ablation of the Di-Long Components.} We investigate the important components of our Di-Long model by starting with a plain architecture with fewer capabilities (without processing semantic maps, with no social information, not using the goal and waypoints, and without distillation) and adding one component at a time. This investigation is reported in Fig~\ref{fig:ablation_dilong}. 
The ablation is conducted on the inD dataset on the long-term prediction setting, and we report the ADE (left) and FDE metrics (right).
We compare our increasingly improving Di-Long model with Goal-SAR and Y-Net. As shown in the table, the use of the semantic maps, the goal and waypoint, and the social information are important, but clearly, our distillation strategy (in different forms) makes a big difference and improves the ADE and FDE, surpassing previous works. 

\begin{figure}[ht]
    \centering
    \includegraphics[width=0.48\textwidth]{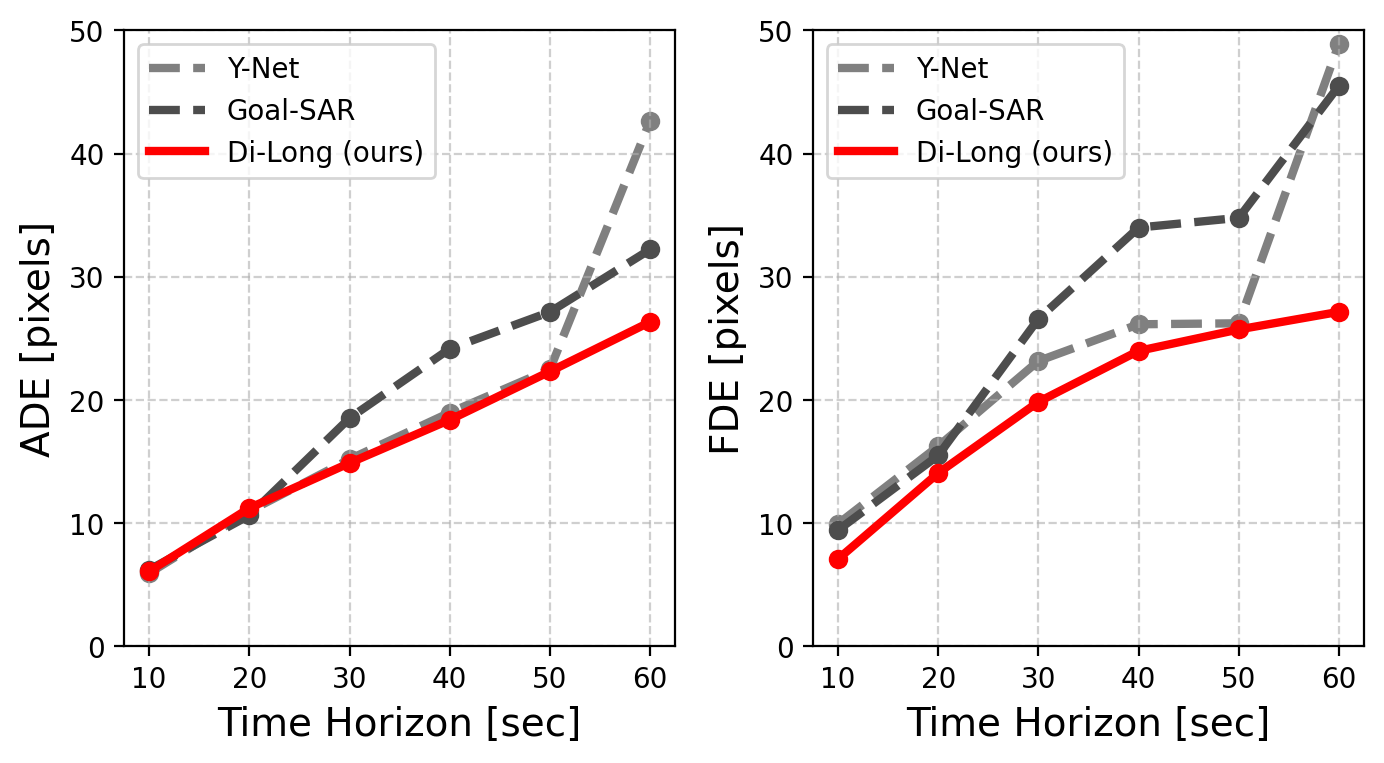}    
    \caption{\small{\textbf{Performance Across Longer Time Horizons.} These results are obtained on the inD dataset.}}
    \label{fig:benchmarking_lt}
\end{figure}

\vspace{2mm}
\noindent
\textbf{Increasing the Long-Term Time Horizon.} 
In Fig.~\ref{fig:benchmarking_lt}, we studied the prediction capability of the Di-Long model on longer time horizons than $t_f = 30$ sec, and compared to Goal-SAR and Y-Net. Specifically, we increase the prediction length starting from $t_f = 10$ sec up to $t_f = 60$ sec.
As shown in the figure, our Di-Long model consistently outperforms Y-Net and Goal-SAR, and interestingly, at shorter time horizons, the performances of the models are comparable, however, when the prediction length increases, the Di-Long model maintains a low degradation, outperforming previous works in terms of ADE and FDE, without manifesting sudden degradation after a critical time horizon. 
\begin{figure}[ht]
    \centering
    \includegraphics[width=0.48\textwidth]{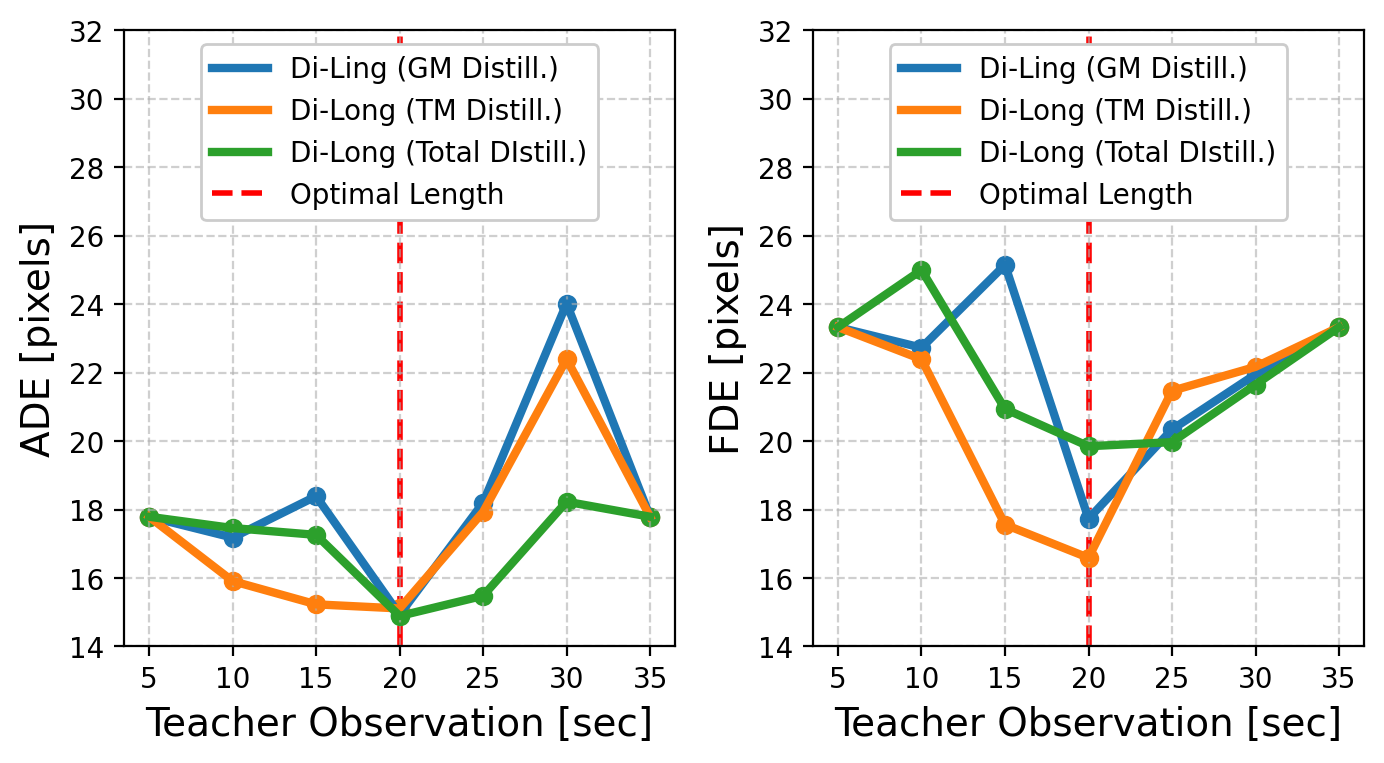}
    \caption{\small{\textbf{Optimal Teacher Observation Length.} These results are obtained on the inD dataset.}}
    \label{fig:optimum_achor}
\end{figure}

\vspace{2mm}
\noindent
\textbf{Optimal Teacher Observation Length.} 
In Fig.~\ref{fig:optimum_achor}, we investigate on the long-term setting on the inD dataset, the impact of the length of the teacher observation $t_p + t_a$ where $t_p$ is the student observation duration, and $t_a$ is the anchor time that is additionally seen by the teacher during training. As shown in the figure, spanning $t_p + t_a$ from $5$ to $35$ the results of the student network behave differently: at the extremes, when the teacher observation is too short or too long, the final results in terms of ADE and FDE are not optimal, and at $t_p + t_a = 20$ sec, there is a minimum in both metrics, suggesting that setting the anchor time to $t_a = 15$ leads to best final results. This value of the anchor interestingly coincides with the time step where we extracted the waypoint from the goal module.

\subsection{Technical Details}
The scene $\mathcal{I}$ can have different dimensions in the considered datasets, and for this reason, following~\cite{Mangalam2020FromGW, Chiara2022GoaldrivenSR}, we resize them to $H = W = 256$, preserving the scale-ratio. 
The considered classes for the semantic segmentation maps $\mathbf{M}$ are $C=\{$\texttt{pavement, terrain, structure, tree, road, not defined}$\}$. 
The U-Nets in the student and the teacher goal estimator have $5$ down-sampling and up-sampling residual blocks, with respectively $[32, 32, 64, 64, 64]$ and $[64, 64, 64, 32, 32]$ channels for the encoder and decoder, as in~\cite{Mangalam2020FromGW}. 
Estimated goals are sampled from $\mathbf{z}$ and re-encoded into $\mathbf{\hat{z}}$, and for sampling, we use the Test-Time-Sampling-Trick \texttt{TTST} proposed in~\cite{Mangalam2020FromGW}. 
We implemented our codebase using PyTorch, and we trained the Di-Long model on a single GeForce RTX $2080$ Ti GPU for $500$ epochs (that corresponds to approximately $21$ hours of training) using the Adam~\cite{Kingma2014AdamAM} optimizer with a batch size of $8$ and a constant learning rate set to $1\mathrm{e}{-4}$. The segmentation network $\mathcal{Q}$ is frozen and initialized with the pre-trained weights from ImageNet-1k, similar to~\cite{Mangalam2020FromGW}.

\subsection{Qualitative Results}
\vspace{-5pt}
\begin{figure}[ht]
    \centering
    \includegraphics[width=0.48\textwidth]{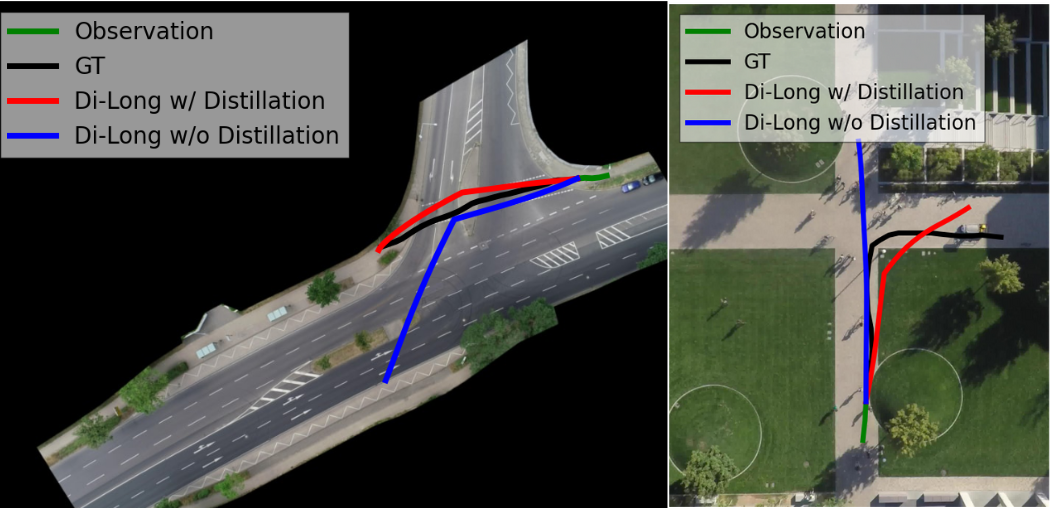}
    \caption{\small{\textbf{Qualitative Results.} Observed trajectories are depicted in green; ground truth (GT) predictions in black; Di-Long without knowledge distillation in blue; Di-Long ``full model'' predictions in red. The left figure shows a long-term prediction example from inD (scene 00), whereas the right one is for long-term prediction in SDD (Hyang0 scene).}}
    \label{fig:ind_sdd_qual}
\end{figure}

In Fig.~\ref{fig:ind_sdd_qual}~(\emph{left}), we report an example where the Di-Long w/o Knowledge Distillation tends to strictly follow the road semantics. 
Still, the final goal is far from the ground truth, whereas due to goal module distillation, the full model predicts the final goal very close to the ground truth. 
In Fig.~\ref{fig:ind_sdd_qual}~(\emph{right}), we show another example in which the Di-Long full model clearly matches the ground truth behavior.

\vspace{-0.5mm}
\section{Conclusions}
In this work, we introduced Di-Long, which is a simple, scene and socially compliant auto-regressive transformer-based architecture empowered with a novel short-to-long term knowledge distillation scheme for trajectory prediction. 
Di-Long achieves state-of-the-art performances in long-term as well as in short-term prediction on both the Intersection Drone (inD) and Stanford Drone (SDD) datasets, thus proving the effectiveness of the proposed solution. 







\bibliographystyle{IEEEtran}
\bibliography{refs} 
\end{document}